# A Retinal Image Enhancement Technique for Blood Vessel Segmentation Algorithm

A. M. R. R. Bandara
University of Moratuwa, Katubedda,
Moratuwa, Sri Lanka.
ravimalb@uom.lk

P. W. G. R. M. P. B. Giragama
Base Hospital Kebithigollawa
Kebithigollawa, Sri Lanka.
mgiracn@yahoo.com

# A Retinal Image Enhancement Technique for Blood Vessel Segmentation Algorithm

A. M. R. R. Bandara
University of Moratuwa, Katubedda,
Moratuwa, Sri Lanka.
ravimalb@uom.lk

P. W. G. R. M. P. B. Giragama
Base Hospital Kebithigollawa
Kebithigollawa, Sri Lanka.
mgiracn@yahoo.com

*Abstract*— The morphology of blood vessels in retinal fundus images is an important indicator of diseases like glaucoma, hypertension and diabetic retinopathy. The accuracy of retinal blood vessels segmentation affects the quality of retinal image analysis which is used in diagnosis methods in modern ophthalmology. Contrast enhancement is one of the crucial steps in any of retinal blood vessel segmentation approaches. The reliability of the segmentation depends on the consistency of the contrast over the image. This paper presents an assessment of the suitability of a recently invented spatially adaptive contrast enhancement technique for enhancing retinal fundus images for blood vessel segmentation. The enhancement technique was integrated with a variant of Tyler Coye algorithm, which has been improved with Hough line transformation based vessel reconstruction method. The proposed approach was evaluated on two public datasets STARE and DRIVE. The assessment was done by comparing the segmentation performance with five widely used contrast enhancement techniques based on wavelet transform, contrast limited histogram equalization, local normalization, linear un-sharp masking and contourlet transform. The results revealed that the assessed enhancement technique is well suited for the application and also outperforms all compared techniques.

## I. INTRODUCTION

Early diagnosis is crucial in many sight-threatening diseases like glaucoma, hypertension and diabetic retinopathy which cause blindness among working age people [1], [2]. Therefore retinal image analysis has become one major diagnosis method in modern ophthalmology. Retinal image analysis typically involves in blood vessel segmentation, optical disc segmentation and fovea segmentation for detecting and analyzing any abnormalities [3], [4]. The contrast enhancement is one mandatory step in any of the related image analysis approaches [1]–[6]. The main challenge of any contrast enhancement algorithm is finding a method to regulate the amplification according to the illumination variations over the image [7]. A typical solution is applying a homomorphic filter to normalize the illumination. However, some contrast enhancement techniques such as contrast limited adaptive histogram equalization (CLAHE) [8] and local normalization (LN)[9] have the capability of analyzing the local illumination and regulate the amplification to bring the final outcome up to an acceptable level of quality. CLAHE is able to handle the illumination variation by doing local histogram equalization and also can regulate the amplification of the details. However, it introduces a box-shaped artifact which may cause to suppress some details and also it amplifies some undesirable details [10].

Speeded up adaptive contrast enhancement (SUACE) algorithm is an outcome of one of our previous studies, in which the goal is about enhancing superficial vein images that have been captured with infrared radiation in real-time [10]. Although SUACE was invented for the purpose of real-time execution on resource constrained devices, it also exhibited several other superior properties such as self-illumination balancing, ability to enhance a particular type of detail and the ability to regulate the detail amplification [10]. In this study, we assess the suitability of SUACE in retinal image enhancement for segmenting the blood vessels.

The evaluation of the suitability of the enhancement technique which is done by human experts is subjective. Therefore we used an existing blood vessel segmentation method combining with SUACE to obtain the accuracy of segmentation, which then used as a measurement to quantify the suitability of the contrast enhancement technique for the blood vessel segmentation task. We used Tyler Coye algorithm [11] which has been used in numerous related studies published recently[1], [3], [6], [12], with a slight modification and combining with SUACE. The modification has been done for better dealing with the noises caused by the contrast enhancement techniques without losing parts of vessels which are detached from the main vessel network due to foreground misclassification. Precisely a connected component analysis based noise removal technique and vessel reconstruction with probabilistic Hough line transformation[13] are used instead of using only connected component based technique which is specified in the original implementation of the Tyler Coye algorithm. In the experiment, the effect of SUACE in blood vessel segmentation is compared by applying the modified Tyler Coye algorithm on images from two public retinal image datasets STARE[14] and DRIVE[15], which were enhanced by different contrast enhancement techniques based on wavelet transform[16], CLAHE, LN, linear unsharp masking (LUM) and contourlet transform[5]. The details in-depth of SUACE algorithm is specified in [10]. However, the algorithm has been briefly described in the next section in order to self-contain the article.

## II. Related Works

### A. SUACE Algorithm

SUACE is based on the linear contrast stretching to the image by selecting the reference dynamic range based on the illumination response around the point which is to be enhanced. The illumination response is obtained by using a low-pass filter with a controllable blur radius. SUACE uses Gaussian smooth filter to obtain the lower frequency band. The two boundaries of the reference range can be calculated by

$$a(x,y) = g(x,y) - \frac{d}{2}$$
$$b(x,y) = g(x,y) + \frac{d}{2} \quad (1)$$

where $a(x,y)$ and $b(x,y)$ are the lower and upper boundaries of the reference range for the scaling, respectively and $d$ is the width of reference dynamic range which is an application dependent variable. The effect of the $d$ has been explained in [10].

The $g(x,y)$ is the low-frequency response of the image which is obtained by filtering the original signal $I(x,y)$ with the Gaussian kernel as shown in

$$g(x,y) = I(x,y) * f(x,y)$$
$$f(x,y) = \frac{1}{2\pi\sigma^2} e^{-\frac{x^2+y^2}{2\sigma^2}} \quad (2)$$

The details about parameters selection have been described in [10]. The enhanced image can be obtained by using

$$\hat{I}(x,y) = \begin{cases} 0 & I(x,y) < a(x,y) \\ 1 & I(x,y) \geq b(x,y) \\ \frac{(I(x,y) - a(x,y))}{d} \times k, & \text{otherwise} \end{cases} \quad (3)$$

where $\hat{I}(x,y)$ is the contrast enhanced image and $k$ is the new dynamic range. Then the enhanced image can be processed with Tyler Coye algorithm.

### B. Tyler Coye Algorithm

Fig. 1 shows the flow chart of the Tyler Coye algorithm for blood vessel segmentation in retinal images. The principle component analysis (PCA) of weighted Lab color model is used for converting the image into grayscale. The contrast enhancement is done by adaptive histogram equalization. Following the contrast enhancement step, it excludes the background by subtracting the average filtered image. Isodata[17] is used for extracting a fair threshold level for the binarization process and finally, the smaller components are removed by considering the size of each connected components. The method of combining the SUACE with Tyler Coye algorithm and the modification done for the original algorithm is described in the next section.

## III. Integration of SUACE

The SUACE algorithm is injected into the process at the contrast enhancement step. The input for the SUACE is the grayscale image which is obtained by PCA based Lab color to grayscale conversion.

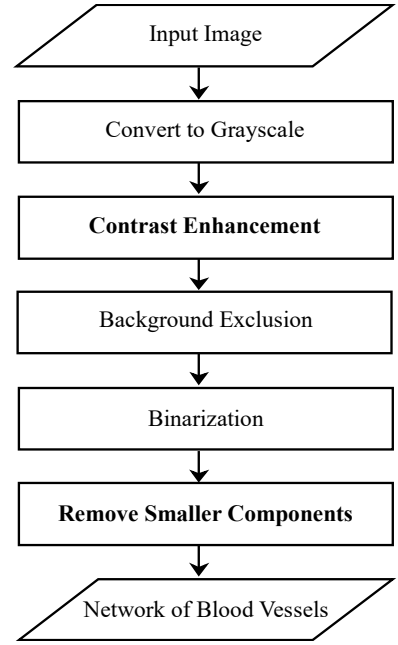

Fig. 1. Flow of Tyler Coye algorithm

It is common that any contrast enhancement algorithm amplifies both the signal and noises simultaneously. In this application, the amplified noises appeared as several connected components which are detached from the network of blood vessels. The simple connected component analysis method which is used in the original implementation of Tyler Coye tends to remove parts of some smaller vessels with the width of a single pixel, due to the discontinuities over the vessel network as shown in Fig. 2. We extend this original noise removal technique by fixing the discontinuities in the vessel network. The specific procedure is given below.

*1)* Remove the connected components with the area less than $a_1$. The value of $a_1$ is selected to be a smaller value somewhere between 5-15 pixels.

*2)* Obtain the probabilistic Hough line transformation of every local window with the size $h$ and fill the disconnected gaps in the vessel network by drawing lines for the each pair of endpoints having the line votes more than the given threshold $v$.

*3)* Remove the connected components with the area less than $a_2$ where $a_2 >> a_1$.

The $a_1$ was selected as it does not remove any of the vessels in the image hence the first step can remove some of the noises safely. The window size $h$ was selected based on the average distance of the discontinued spaces in the vessel networks. Since the window is small, the threshold for the Hough line votes should be kept smaller. We empirically select a value for $v$ throughout the experiments. After fixing the discontinuities, only the noises are remained without connecting to the vessel network. Therefore we can set a larger value to $a_2$ in the third step to remove all the noises safely without false classifying the vessels as the background. The Hough line transformation votes for potential lines having similar angles hence a slight discontinuity in a vessel does not affect the voting process. Therefore the discontinuities can be fixed computationally by drawing lines with the highest votes.

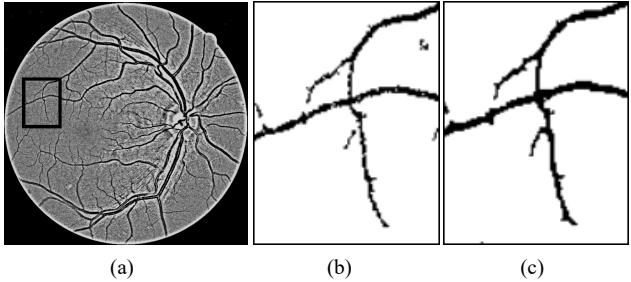

Fig. 2. Hogh line transformation based reconstruction. **(a)** Image with a sample region marked; **(b)** Binarized image of the sample region; **(c)** Image after the noise removed in vessel reconstructed image.

## IV. EXPERIMENTAL SETUP

The suitability of SUACE is evaluated by comparing the segmentation accuracy of blood vessels with other state-of-the-art contrast enhancement techniques. The $\sigma$ and $d$ parameters of the SUACE were set to 7 and 16 respectively throughout the experiments done in this study. These values were selected empirically where the best average performance is given in the experiments. Further we set the first connected component area threshold $a_1$ to 10, the window size $h$ to 5, the Hough line vote threshold $v$ to 3 and the second connected component area threshold $a_2$ to 50 throughout the experiments. We repeated the same procedure for the alternative contrast enhancement algorithms used in this study. The segmentation performance is measured by using the accuracy (ACC) which is defined by the sum of the true positives (pixels correctly classified as on a vessel) and the true negatives (correctly classified as on non-vessel region), divided by the total number of pixels in the images. Further, we obtain the true positive rate (TPR) and false positive rate (FPR) to measure the correctness of the segmentation. TPR is defined as the total number of true positives, divided by the number of true blood vessel pixels. The FPR is calculated as the total number of false positives divided by the number of non-vessel pixels marked in the ground truth image. We report the average of TPR, FPR, and accuracy for two publicly available datasets namely DRIVE and STARE.

## V. RESULTS AND DISCUSSION

Fig. 3 shows the output of a sample image which has been enhanced by each of the contrast enhancement methods. In addition to the blood vessels in the image, there are several other components such as the fovea which is the darker region at the middle of the image and the optical disc which is at the convergence point of all the blood vessels. SUACE has enhanced the blood vessels yet suppressed the fovea and most part of the optical disk. However, all other algorithms except for the one based on the LN have amplified the signal from fovea and the optical disc. But the LN based method has introduced a kind of random noise all over the image hence the enhancement become less usable for the vessel segmentation. Both the contourlet and wavelet based contrast enhancement algorithm have introduced black patches to the image in different scales which can be hard to remove in the segmentation process. Besides these newly introduced noises, the contourlet based enhancement technique suffers from the imbalance illumination problem compared to the uniform illumination distribution of the output of SUACE.

Fig. 3 also shows the results after segmenting the images enhanced by the 6 different contrast enhancement algorithms with their ground truth segmentations. Both the CLAHE and LN based methods show that it detects most part of the vessels and also misclassified most of the background as well. The LUM, wavelet and contourlet based methods show a less misclassification of background yet misclassified several vessel areas. It is clearly observable that the segmentation performance over the image which has been enhanced by SUACE has correctly segmented most of the blood vessels including the smaller vessels.

The result of quantitative analysis of the segmentation is given in Table I. The enhancement of the SUACE outperforms the other methods by means of TPR and accuracy. The reason might be that there is no threshold value which can be used to separate the foreground and background all over the image due to the imbalance illumination produced by the other enhancement methods. The higher threshold provided by isodata algorithm might remove most of the noises as well as the part of the vessels in the network.

The lowest FPR is shown by the wavelet-based method where SUACE shows the second best value. However, the TPR of the wavelet-based method is extremely lower. The fact proves that the SUACE still can compete with the wavelet-based method. The TPR is gained by the contribution of both the Hough line transformation based vessel reconstruction and the performance of the contrast enhancement algorithm. However, due to the vessel reconstruction process is done with all the compared contrast enhancement techniques; we can conclude that the best TPR is achieved due to the relative contribution from the SUACE algorithm. Table I also showed that the integration of the SUACE with the modified Tyler Coye algorithm achieved the best accuracy compared to the other integrated approaches.

TABLE I. PERFORMANCE COMPARISON

| Method | STARE | | | DRIVE | | |
|---|---|---|---|---|---|---|
| | *TPR* | *FPR* | *ACC* | *TPR* | *FPR* | *ACC* |
| *CLAHE* | 0.7765 | 0.0520 | 0.9245 | 0.7414 | 0.0498 | 0.9122 |
| *LUM* | 0.7789 | 0.0326 | 0.9112 | 0.7698 | 0.0331 | 0.9058 |
| *Wavelet* | 0.6017 | **0.0311** | 0.8954 | 0.5958 | **0.0315** | 0.8902 |
| *Contorulet* | 0.6724 | 0.0325 | 0.9012 | 0.6733 | 0.0327 | 0.9101 |
| *LN* | 0.7456 | 0.0942 | 0.8735 | 0.7389 | 0.0955 | 0.8598 |
| *SUACE* | **0.7885** | 0.0324 | **0.9489** | 0.7432 | 0.0334 | **0.9411** |

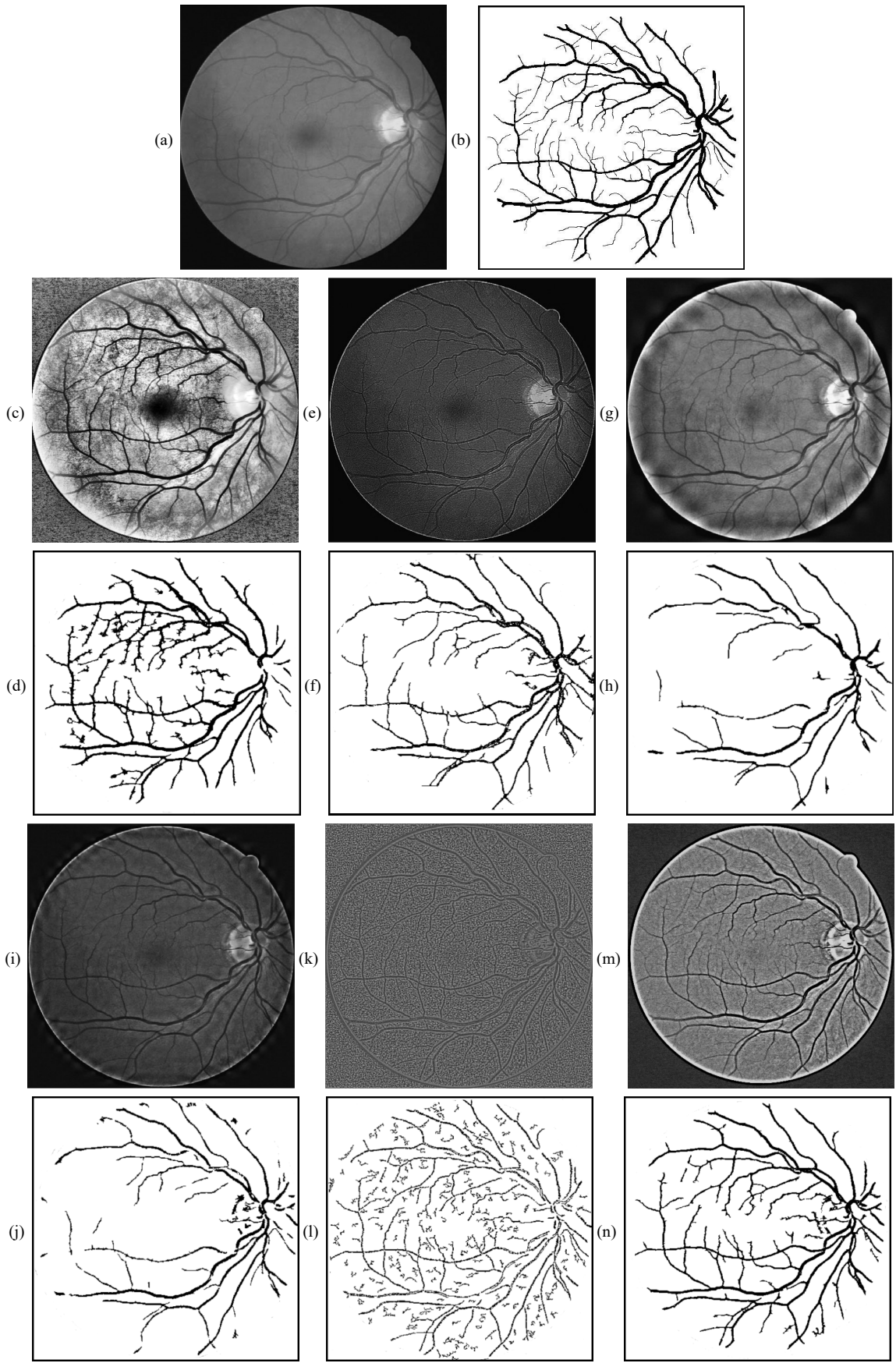

Fig. 3. Outputs of different enhancement methods applied to a sample image from DRIVE dataset. **(a)** & **(b)** Original image and its ground truth segmentation respectively; **(c)** & **(d)** output of CLAHE and its segmentation; **(e)** & **(f)** output of LUM and its segmentation; **(g)** & **(h)** output of Wavelet and its segmentation; **(i)** & **(j)** output of Contourlet and its segmentation; **(k)** & **(l)** output of LN and its segmentation; **(m)** & **(n)** output of SUACE and its segmentation.

## VI. Conclusion

In this paper, we presented an assessment of the suitability of a recently invented contrast enhancement algorithm namely SUACE for retinal blood vessel segmentation in fundus images. The results of both the qualitative and quantitative analysis showed the superiority of SUACE in enhancing retinal images for blood vessel segmentation, hence we can conclude that SUACE is not only suitable for this application but also outperforms all other enhancement techniques which are used in the comparison. Further works can be directed to tune up the parameters of SUACE to detect the other possible components in retinal images such as fovea, optical disc and retinal lesions.